\documentclass[conference,a4paper]{IEEEtran}
\IEEEoverridecommandlockouts

\usepackage[
    colorlinks=true,
    linkcolor=blue,
    citecolor=blue,
    urlcolor=cyan
]{hyperref}

\usepackage[cmex10]{amsmath}
\usepackage{amssymb,amsfonts}
\usepackage{dblfloatfix}

\usepackage[ruled,vlined]{algorithm2e}
\usepackage{graphicx}
\graphicspath{{Figures/PDF/}{Figures/PNG/}}

\usepackage{booktabs}
\usepackage{siunitx}
\usepackage[numbers,compress]{natbib}
\usepackage{texnames}
\usepackage{bm,bbm}
\usepackage{orcidlink}
\usepackage{subcaption}
\usepackage{multirow}
\usepackage{authblk}
\usepackage{xcolor}
\usepackage{pifont}
\newcommand{\cmark}{\ding{51}}
\newcommand{\xmark}{\ding{55}}

\begin{document}


\title{
\uppercase{Towards a satellite image manipulation and deepfake localization benchmark dataset}}
\author[1]{Jacob Arndt\orcidlink{0000-0002-1097-0428}}
\author[1]{Debvrat Varshney\orcidlink{0000-0001-8898-1736}}
\author[1]{Philipe Dias\orcidlink{0000-0001-9427-7112}}
\author[1]{Nivedita Nukavarapu\orcidlink{0000-0001-9736-916X}}
\affil[1]{\small{\textit{Oak Ridge National Laboratory}\\Oak Ridge, TN, USA\\
\{arndtjw, varshneyd, ambroziodiap, nukavarapun\}@ornl.gov}}

\maketitle
\begin{abstract}
Verifying the authenticity of satellite imagery has become increasingly critical given advances in generative artificial intelligence. Highly realistic synthetic imagery produced for malicious purposes (deepfakes) can have major consequences in the remote sensing domain,  where this data is a fundamental source of information for science applications, planning, logistics, and monitoring. 
The remote sensing community lacks high-quality, fine-grained manipulation datasets suitable for training and evaluating detection and image forensics algorithms. Existing datasets are lacking and those that do exist either provide no ground truth masks for evaluating manipulation localization, or consist of entire images generated by GANs or diffusion models, which are inadequate for measuring localization performance. To address this gap, we describe a preliminary dataset construction process and prototype benchmark dataset for satellite image manipulation detection and localization. 
The dataset contains 60 images total, with 30 images carefully manipulated using three manipulation types including copy-paste splicing and diffusion model inpainting, and 30 authentic images. Each image is accompanied by a ground-truth mask and acquisition metadata, enabling both pixel-level localization metrics, image metadata studies, and analyses of how manipulation detection performance relates to image collection parameters. We describe the dataset construction process and present this initial release to support further research in image forensics and geospatial deepfake detection. The prototype dataset can be downloaded at \url{https://huggingface.co/datasets/geodf/fmow-fake-small}.
\end{abstract}


\section{Introduction}

Advancements in generative models have made image editing and image synthesis increasingly accessible and highly realistic. While these advancements have benefits ranging from creative applications to generating synthetic data for improving models, they also raise concerns regarding image authenticity and misinformation. As image manipulation and synthesis becomes more sophisticated and realistic, there is a critical need for benchmark deepfake datasets that capture the diversity and realism of traditional and modern image manipulation techniques.

\begin{figure}[t]
\centering
    \begin{subfigure}{\linewidth}
        \includegraphics[width=\linewidth]{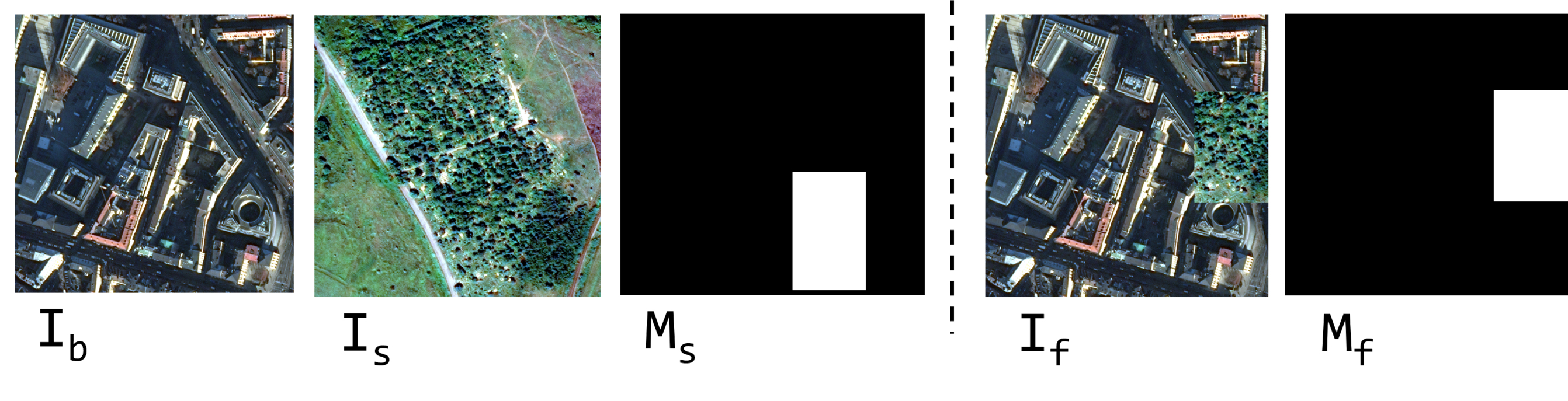}
        \caption{Simple splice workflow randomly crops a source image and pastes the crop into the base image at a random location.}
    \end{subfigure}
    
    \begin{subfigure}{\linewidth}
        \includegraphics[width=\linewidth]{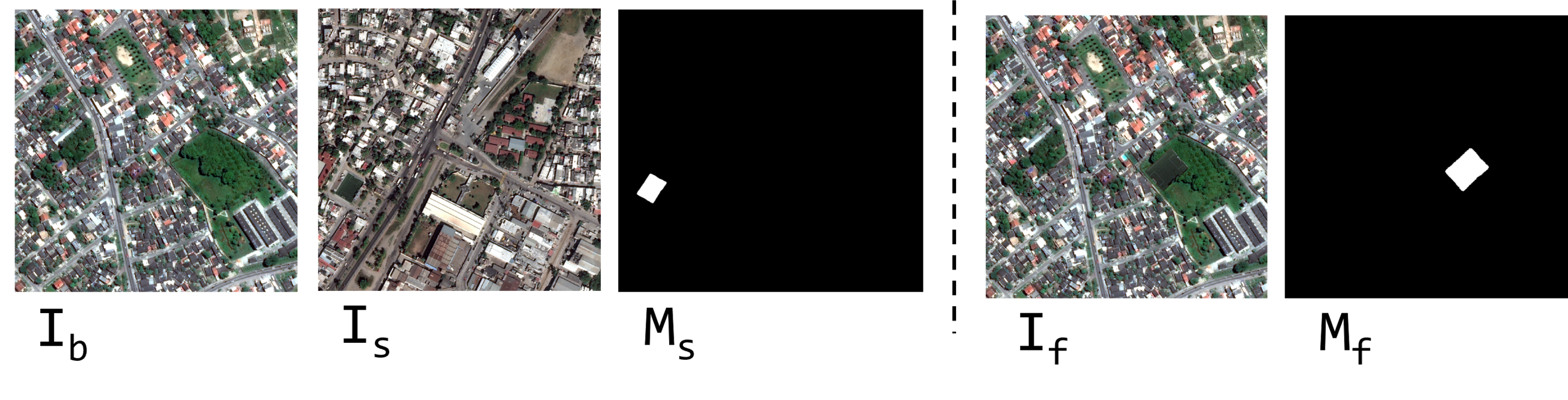}
        \caption{Object splice workflow crops specific objects from a source image and pastes them into a base image at a sensible location.}
    \end{subfigure}
    
    \begin{subfigure}{\linewidth}
        \includegraphics[width=\linewidth]{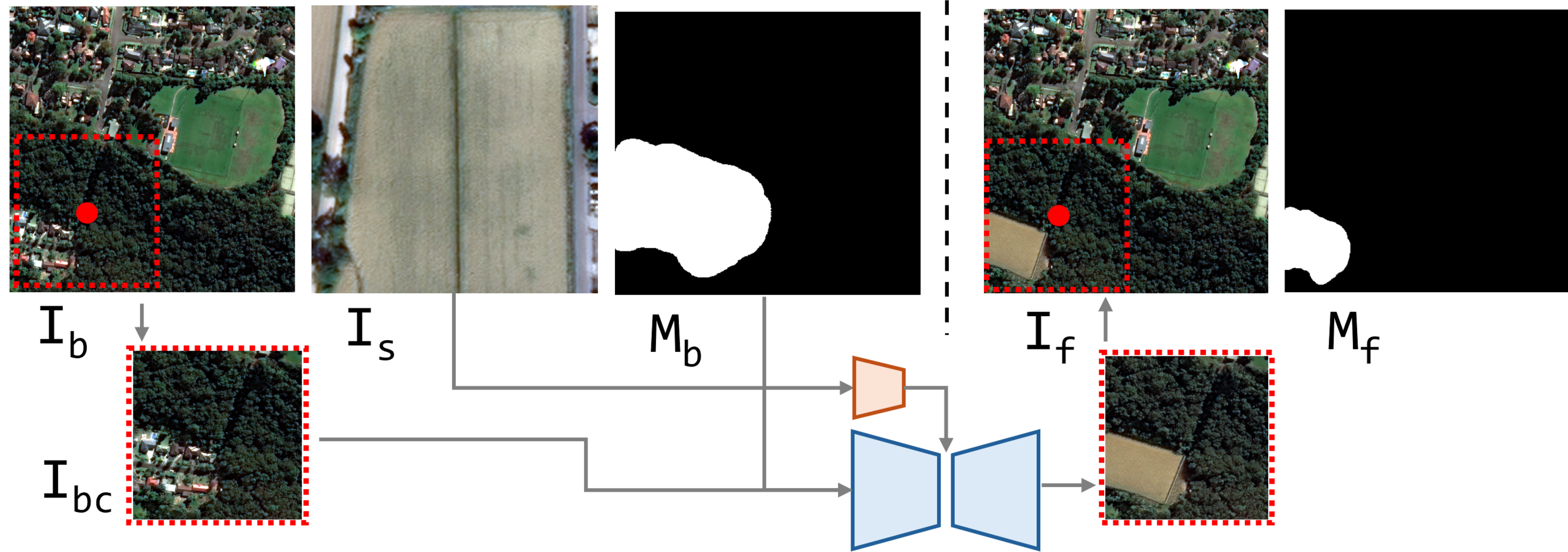}
        \caption{Diffusion inpainting workflow inpaints content from the source image into the base image at some predefined location defined by a base mask using a finetuned diffusion model.}
    \end{subfigure}
\caption{Illustrated workflows for creating the fmow-fake-small dataset. We include three types of fake images in the dataset and each sample in the dataset consists of an RGB image and ground truth mask indicating the manipulated region.}
\label{fig:workflow}
\end{figure}

\begin{table*}[t]
    \centering
    \begin{tabular}{lccccccc}
        \toprule
        \multirow{2}{*}{\textbf{Dataset}} &
        \textbf{Num. Images} &
        \multirow{2}{*}{\textbf{Tasks}} & 
        \multirow{2}{*}{\textbf{Manipulations}} &
        \textbf{Ground Truth} & \multirow{2}{*}{\textbf{Georeferenced}} &\textbf{Acquisition} \\
        &  \textbf{(Real/Fake)} & & & \textbf{Masks} & & \textbf{Metadata}\\ 
        \midrule
        RSFAKE-1M \cite{tan2025rsfake} & 500k/500k & cls. & diffusion synthesis, diffusion inpainting & \xmark & \xmark  & \xmark \\
        FLDCF \cite{sui2024fldcf} & 2846/8350 & cls., seg. & diffusion inpainting (RePaint, LaMa), splicing & \cmark &  \xmark & \xmark \\
        DM-AER \cite{dmaer} & 140k/1m & cls. & StyleGAN2 & \xmark & \xmark  & \xmark \\
        FSI \cite{zhao2021deep} & 4032/4032 & cls. & CycleGAN & \xmark & \xmark  & \xmark \\
        \midrule
        fmow-fake-small & 30/30 & cls., seg. & splicing, diffusion inpainting & \cmark & \cmark & \cmark \\
        \bottomrule
    \end{tabular}
    \caption{Remote sensing deepfake and image manipulation datasets. (cls.: classification, seg.: segmentation)}
    \label{tab:related-datasets}
\end{table*}

\begin{figure*}[t]
    \centering
    
    \begin{subfigure}{0.48\textwidth}
        \centering
        \includegraphics[width=0.3\linewidth]{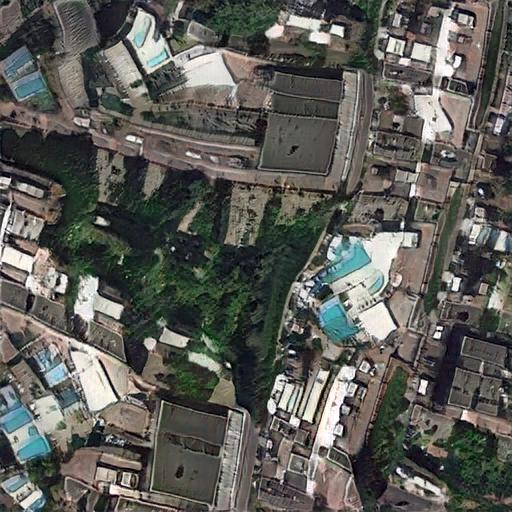}
        \includegraphics[width=0.3\linewidth]{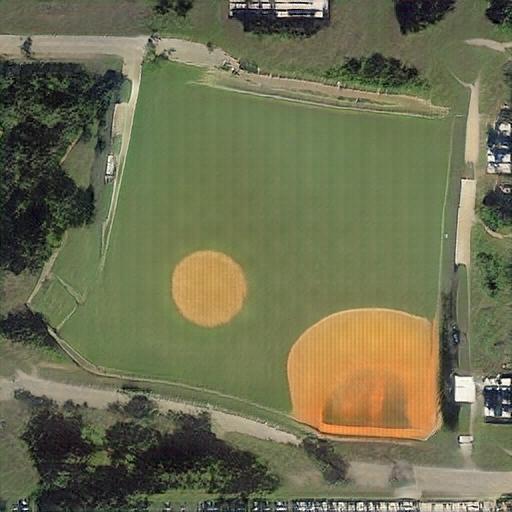}
        \includegraphics[width=0.3\linewidth]{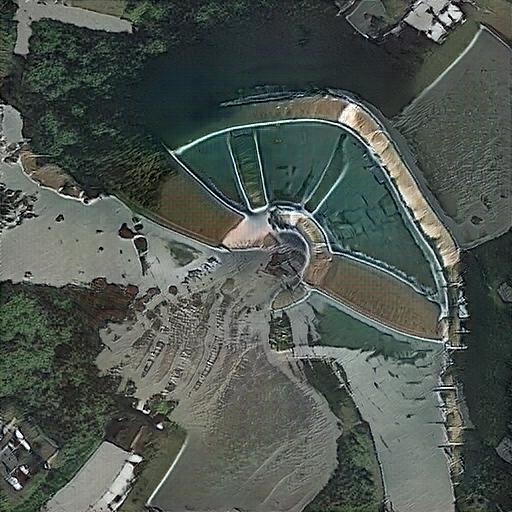}  
        \caption{DM-AER \cite{dmaer}}
        \label{fig:qual-dmaer}
    \end{subfigure}
    \begin{subfigure}{0.48\textwidth}
        \centering
        \includegraphics[width=0.3\linewidth]{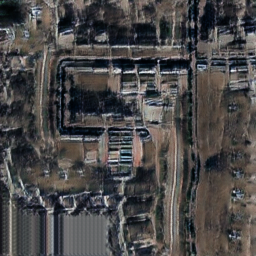}
        \includegraphics[width=0.3\linewidth]{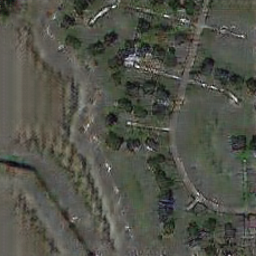}
        \includegraphics[width=0.3\linewidth]{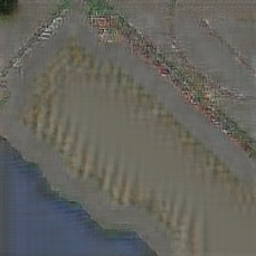} 
        \caption{FSI \cite{zhao2021deep}}
        \label{fig:qual-fsi}
    \end{subfigure}

    \vspace{0.3cm} 

    \begin{subfigure}{0.48\textwidth}
        \centering
        \includegraphics[width=0.3\linewidth]{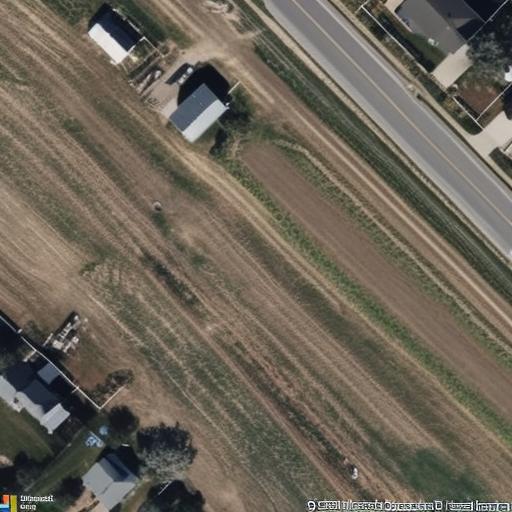}
        \includegraphics[width=0.3\linewidth]{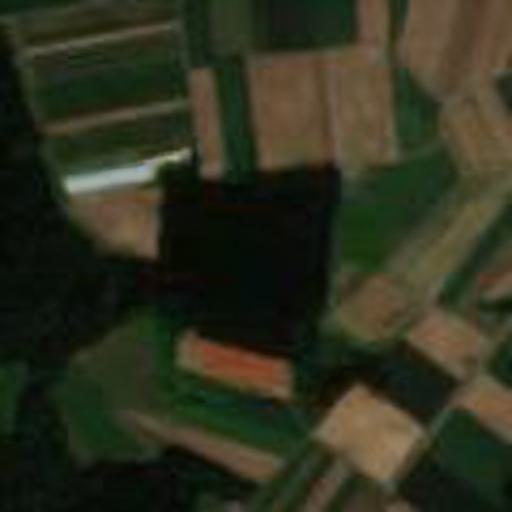}
        \includegraphics[width=0.3\linewidth]{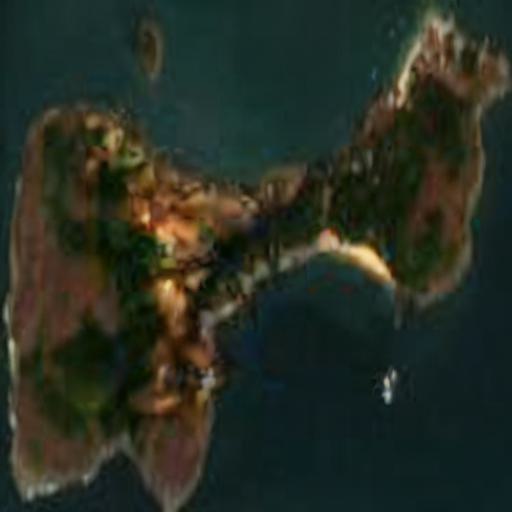} 
        \caption{RSFAKE-1M\cite{tan2025rsfake}}
        \label{fig:qual-rsfake}
    \end{subfigure}
    \begin{subfigure}{0.48\textwidth}
        \centering
        \includegraphics[width=0.3\linewidth]{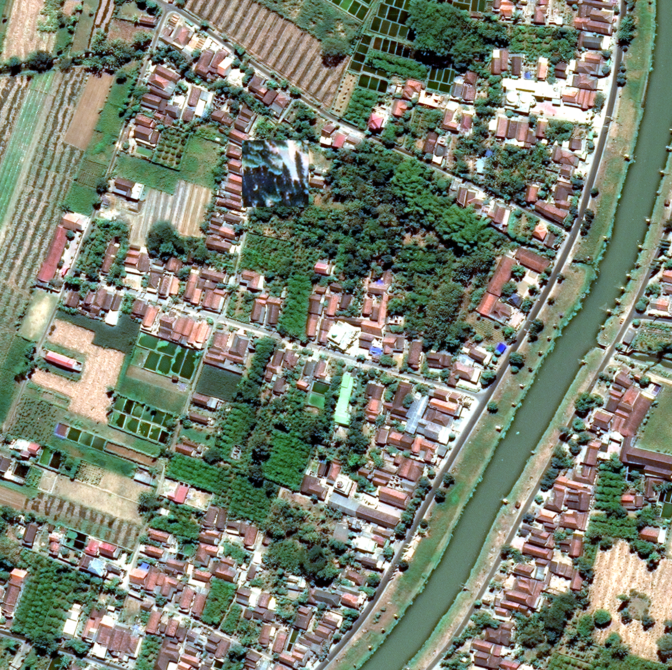}
        \includegraphics[width=0.3\linewidth, height=2.62cm]{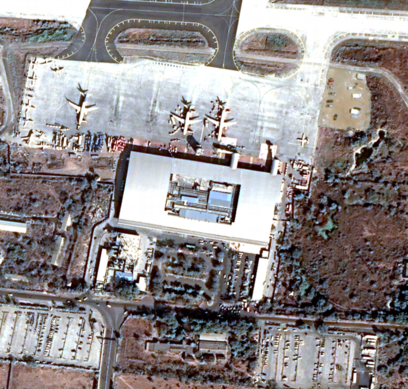}
        \includegraphics[width=0.3\linewidth]{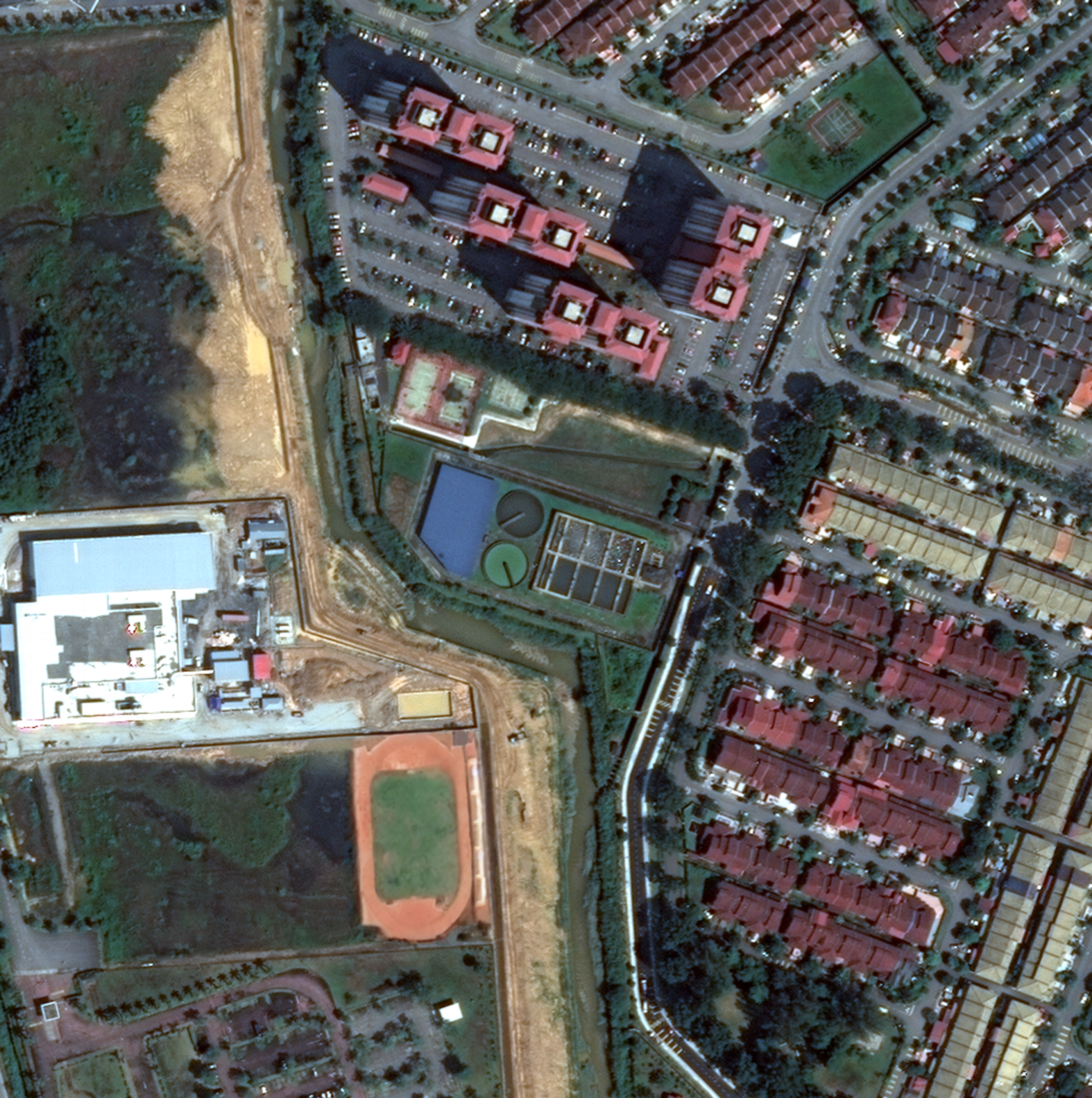}    
        \caption{fmow-fake-small (ours)}
        \label{fig:proposed}
    \end{subfigure}

        \caption{A comparison of samples in existing benchmark fake geospatial datasets (a,b,c) and our proposed dataset (d). Noticeable artifacts in current benchmark datasets such as blurriness, low-resolution, warped features, and web-logos give a clear, easy indication that the images are fake. (d) represents, from left to right, a simple-splice, object-splice, and an inpainted image in the proposed fmow-fake-small dataset.}
\label{fig:datasets_comparison}
\end{figure*}

Current remote sensing deepfake datasets, summarized in Table~\ref{tab:related-datasets}, lack high-quality synthetic examples. As illustrated in Figure~\ref{fig:datasets_comparison} many samples in these datasets contain visual artifacts resulting in fake examples that do not fully challenge human perception. Similar observations were made in \cite{yan2025a} in reference to large-scale non-remote sensing deepfake benchmark datasets. In \cite{dmaer, zhao2021deep}, generative adversarial networks (GANs) were used, making the datasets outdated relative to current state-of-the-art generative image models. In \cite{yarlagadda2018satellite}, a dataset consisting of object splices and masks is presented and used in experiments, however the dataset is unavailable and the details presented in the paper are cursory and incompatible with the visual examples. While recent datasets \cite{tan2025rsfake, sui2024fldcf} include examples generated by diffusion models, their use in evaluating deepfake localization is problematic. For RSFAKE-1M \cite{tan2025rsfake}, the inpainting examples lack ground truth masks, 
while for FLDCF \cite{sui2024fldcf}, the inpainting examples are intended to be exact replicas of their ground truth images using RePaint and LaMa.

In this paper, we propose a new satellite image forgery and manipulation dataset suitable for evaluating image manipulation and deepfake localization. The manipulations in the dataset range in complexity, visual perceptual quality, and size presenting an opportunity for thorough analysis of failure and success cases for detection methods. Three types of manipulations are present in the dataset: i) simple random copy-paste splices; ii) object copy-paste splices; and iii) diffusion model inpainting examples. Figure~\ref{fig:workflow} illustrates the dataset construction workflow for each of these manipulations. The dataset is sourced from the Functional Map of the World (fMoW) dataset \cite{christie2018functional} and includes 30 manipulated images and 30 authentic images. Each example includes a three channel (RGB) image, a ground truth mask indicating the manipulated region, and metadata describing the source image collection characteristics and the base image collection characteristics. 

The contributions of this paper are three-fold. First, we construct a realistic georeferenced remote sensing deepfake dataset that includes diverse image manipulations ranging from traditional splices to sophisticated diffusion model inpainting. Second, the dataset includes ground truth masks and extensive metadata for assessing deepfake localization across different manipulation types. We also release our curated reference object images that we used for the object splices and inpainting. Third, using georeferenced imagery, we present an approach for obtaining scale consistent geospatial object inpainting for the exemplar-based image editing setup. The remainder of the paper describes the dataset construction process, followed by results and discussion.

\section{Methods}

\subsection{Preliminaries}
Our dataset is derived from the fMoW \cite{christie2018functional} dataset, a high-resolution satellite image scene classification dataset consisting of over 1 million images collected over approximately 200 countries by WorldView-2, WorldView-3, GeoEye-1 and QuickBird-2 sensors. For the fmow-fake-small dataset, we use the three channel (RGB), 8-bit, pansharpened images in the fMoW training set. The RGB and not the multispectral version of the dataset is used so that pretrained diffusion models can be used for inpainting, which require three-channel inputs. We begin by georeferencing and projecting all images in the dataset using information found in the groundtruth metadata for each image. We use the $mean\_pixel\_width$ and $mean\_pixel\_height$ to set the pixel size, the $utm$ to determine the projected coordinate reference system for projecting the image, and we set the geographic origin of the image by using the $raw\_location$. We save each image as a GeoTIFF.



\subsection{Simple Splice}\label{sec:simple}
We define a simple splice as a random crop copied from a splice source image $I_s$ and pasted into a different base image $I_b$. We define four crop sizes for these simple splices: $16\times16$, $64\times64$, $128\times128$, $256\times256$ pixels. Based on the given crop pixel size, we calculate the corresponding crop size in the base image in physical units (meters), and then use this physical size to crop from the splice source image. To paste the crop into the base image, we resample the crop to match the base image's x- and y- pixel resolution by using $2\times2$ bilinear resampling. This cropping and resampling ensures the pasted crop has the same pixel resolution and spatial extent, thus making the crops between image scales consistent. 

\subsection{Object Splice}
For object splices, we segment ``stuff'' or ``things'' from the splice source image using the segment anything model \cite{kirillov2023segment} and paste the segmented content into a base image at a reasonable location. To ensure such quality, we manually curate the object masks and manually paste them into the base image. Similar to the simple splice portion of the dataset, we create object splices of different sizes by targeting stuff or things ranging from small sizes (e.g. cars, boats, small structures, individual trees), to extra large sizes (e.g. warehouses, major roadways, agriculture fields, stadiums, forests). 

\subsection{Diffusion Model Inpainting}
For inpainting, we perform exemplar-based image editing \cite{yang2023paint} using the RSPaint model developed by \cite{immanuel2025tackling}. The inpainted examples in the RSFAKE-1M dataset \cite{tan2025rsfake} also use this model to generate fake examples. RSPaint is a pretrained Stable Diffusion model \cite{rombach2022high} that has been finetuned on the SAMRS dataset \cite{wang2023samrs} following the training procedure proposed in paint-by-example \cite{yang2023paint}. Three inputs are required for exemplar-based inpainting: a base image $I_b$, a mask $M_b$ indicating the region in the base image to inpaint, and a source image $I_s$ with the content that will be inpainted into the base image. We inpaint content exemplified in a source image into the base image at a location defined by a mask. Limitations of this workflow, as highlighted in \cite{immanuel2025tackling}, include i) the need to manually define the location of the mask for sensible object placement, ii) object scale issues whereby the reference object expands to occupy the entire masked area, and iii) poor inpainting results when the object mask size is not between 15\%-30\% of the image area.

We propose two simple data processing steps to address the second and third challenges highlighted above. Images composing the fMoW dataset range from thousands ($256\times256$) to millions ($4096\times4096$) of pixels and have variable pixel resolution ($0.31$ meters - 1 meter). Given the large image sizes, its difficult to generate high-quality inpainting results for smaller objects which typically only occupy hundreds of pixels in a large image. To address this, we crop the larger image centered on the location of a user defined mask and ensure the crop is large enough such that the object mask is 15\%-30\% of the cropped image area. Second, for cases where we wish to inpaint specific objects (e.g. airplane, car, truck, house, boat) we ensure that the object mask is the same size in physical units (meters) as the object in the reference image. This offers scale consistency, which is particularly important when inpainting objects in remote sensing images. Maintaining the absolute scale of objects and their scale relative to other objects in the base image is crucial for realistic inpainting, because objects such as cars, planes, boats, houses, have specific sizes that must be adhered to in order to appear real. The placement of the inpainted content is determined manually so that content is inpainted at a sensible location. After the inpainted result is generated, we resample the generated content to match the same H$\times$W in pixels as the input to the diffusion model and then crop it back into the larger image using the object mask.


\section{Results}

\begin{figure*}[t]
\centering
    \subcaptionbox{Simple splice examples. left: $16\times16$, center: $128\times128$, right: $256\times256$. \label{fig:row1}}{
        \begin{subfigure}{0.32\textwidth}
            \centering
            \includegraphics[width=\linewidth]{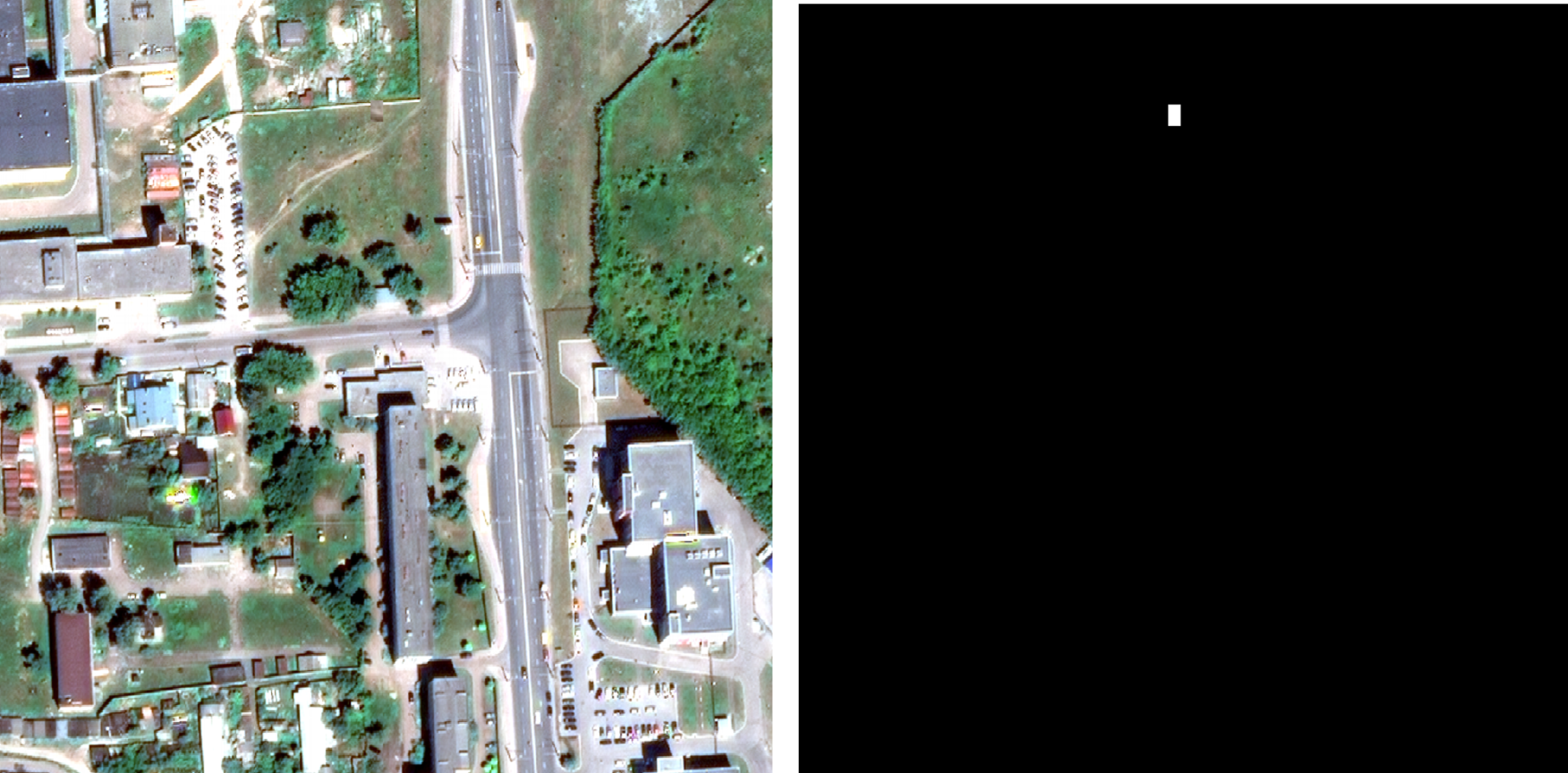}
        \end{subfigure}
        \hfill
        \begin{subfigure}{0.32\textwidth}
            \centering
            \includegraphics[width=\linewidth]{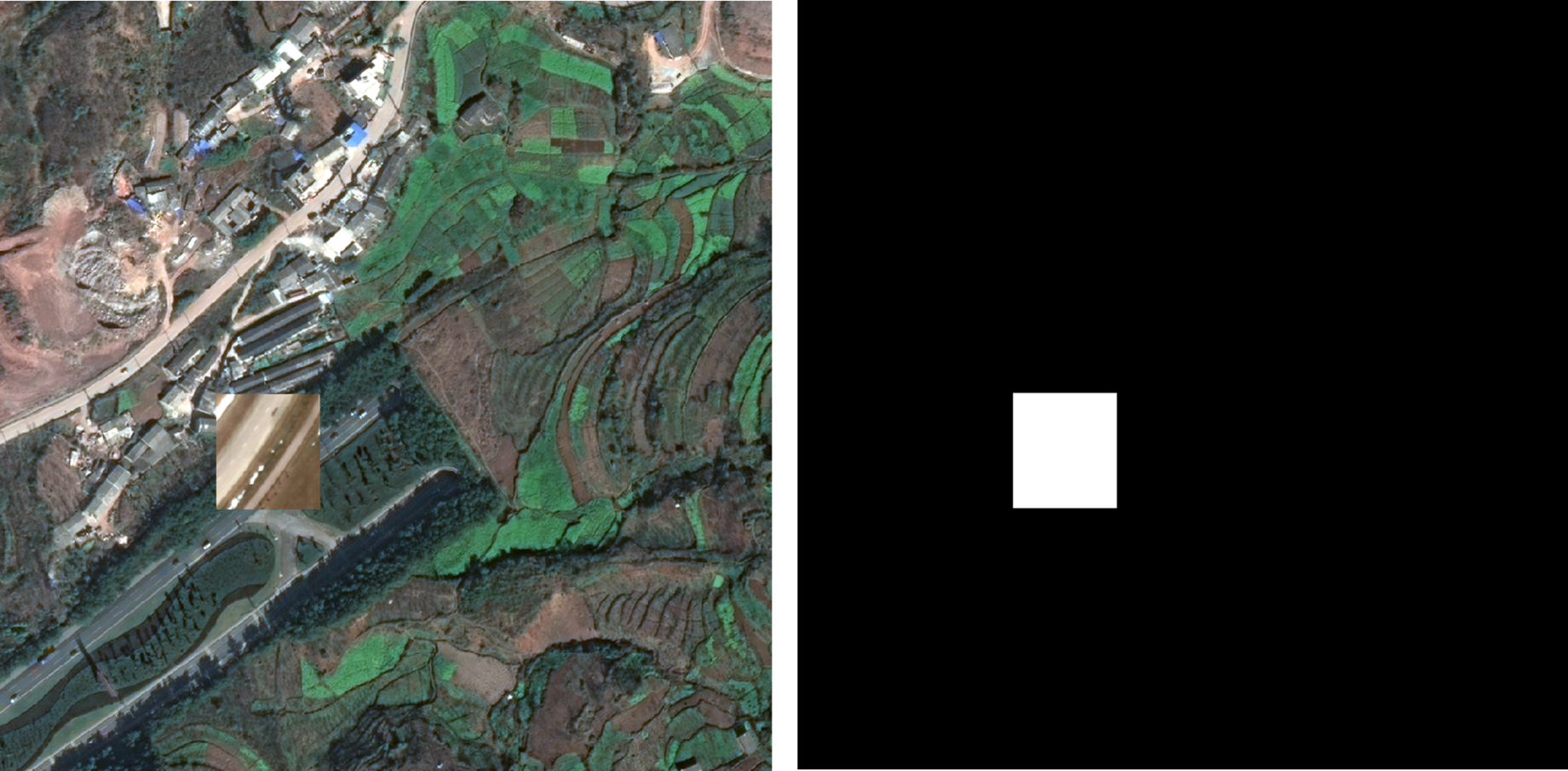}
        \end{subfigure}
        \hfill
        \begin{subfigure}{0.32\textwidth}
            \centering
            \includegraphics[width=\linewidth]{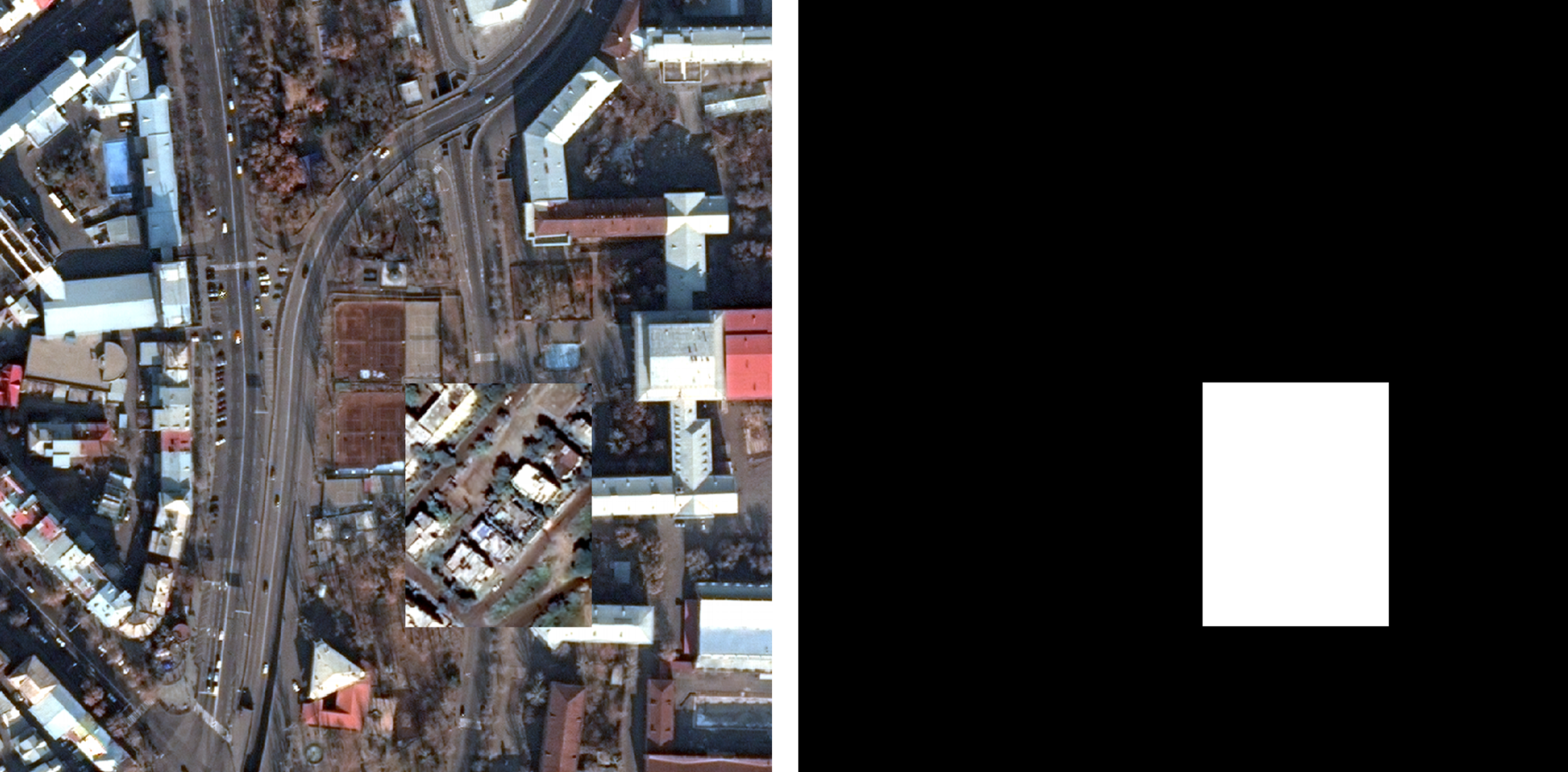}
        \end{subfigure}
    }
    \vspace{0.5em}
    
    \subcaptionbox{Object splice examples. left: car, center: airplane, right: residential block. \label{fig:row2}}{
        \begin{subfigure}{0.32\textwidth}
            \centering
            \includegraphics[width=\linewidth]{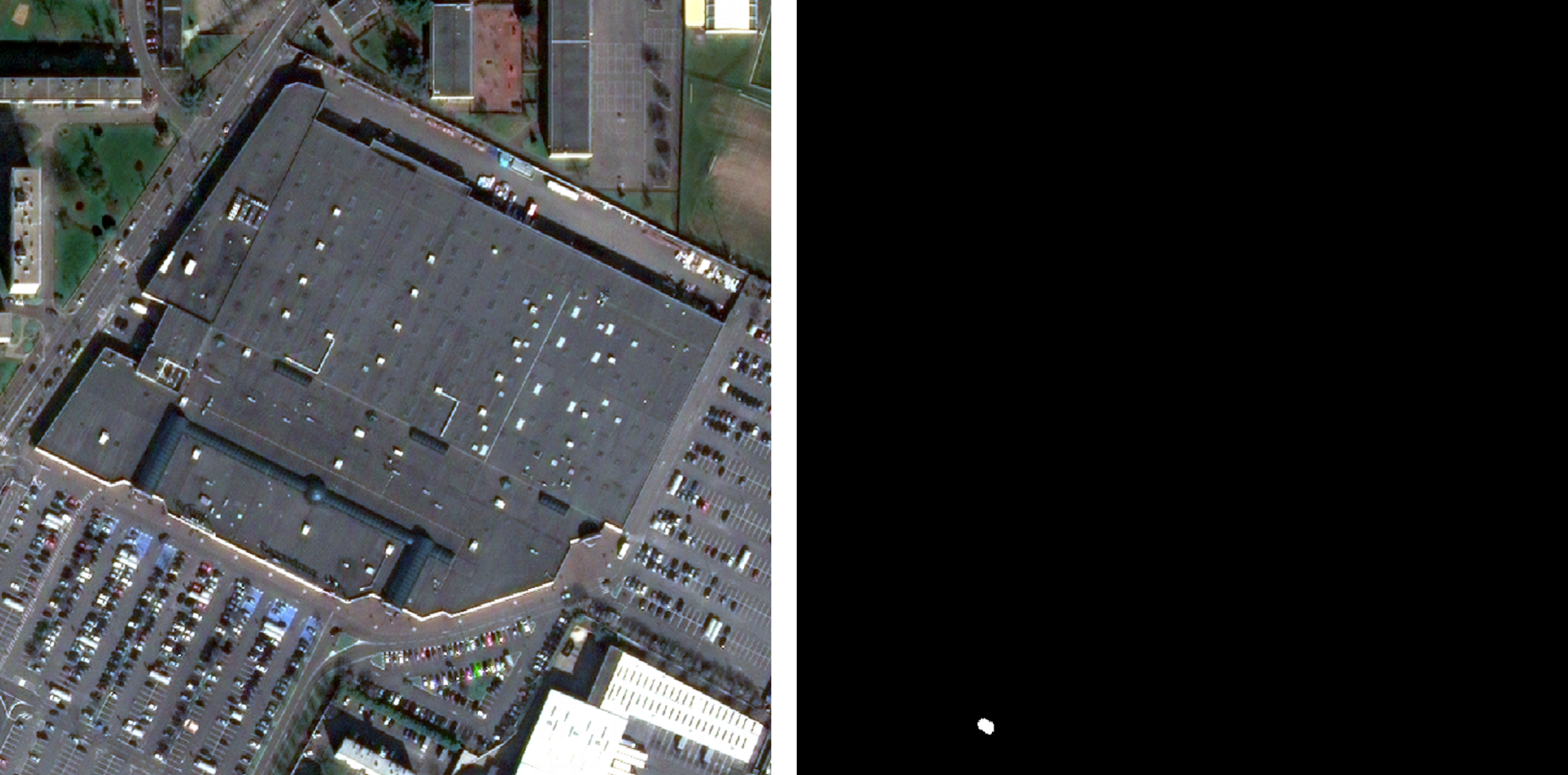}
        \end{subfigure}
        \hfill
        \begin{subfigure}{0.32\textwidth}
            \centering
            \includegraphics[width=\linewidth]{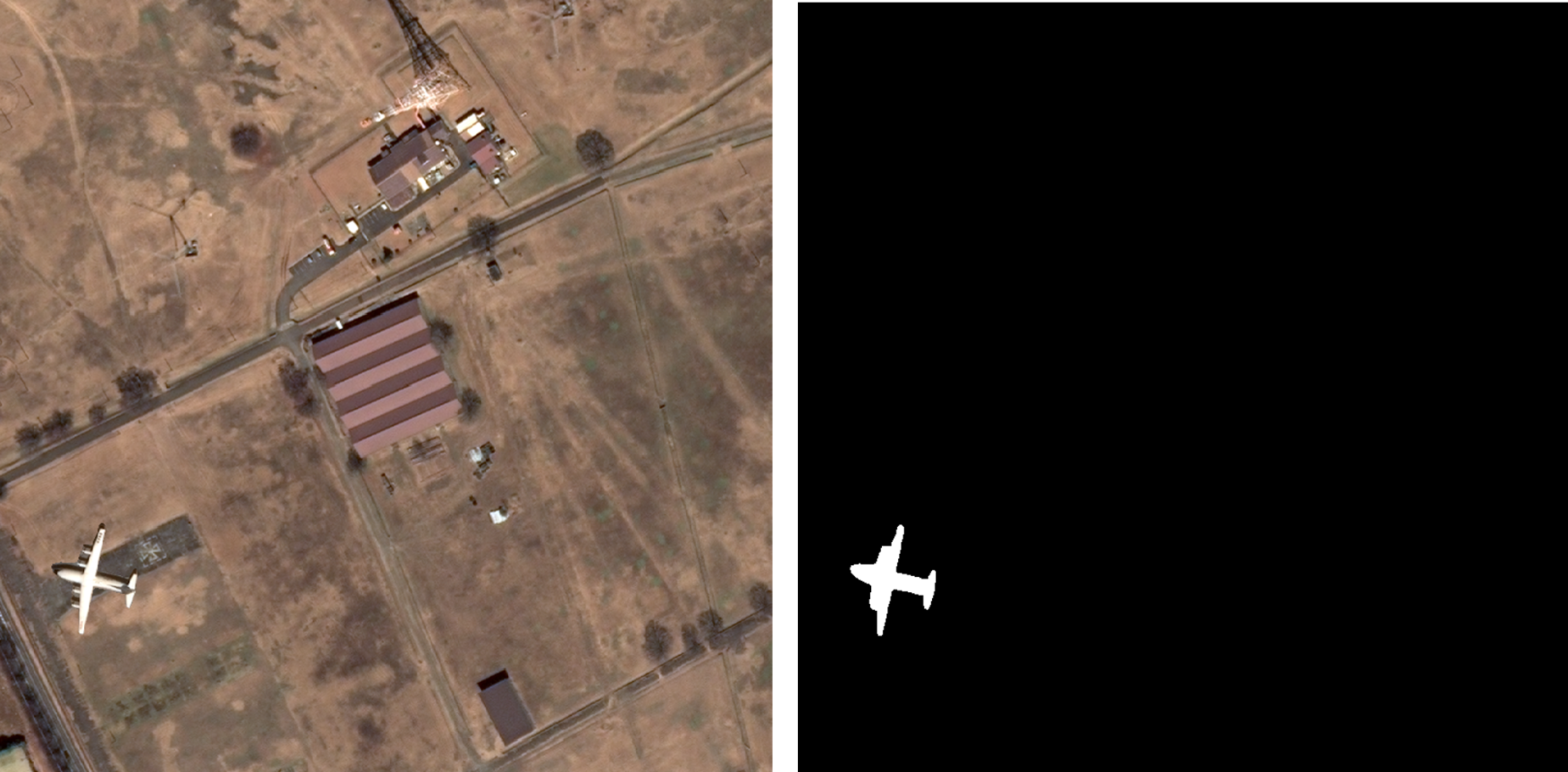}
        \end{subfigure}
        \hfill
        \begin{subfigure}{0.32\textwidth}
            \centering
            \includegraphics[width=\linewidth]{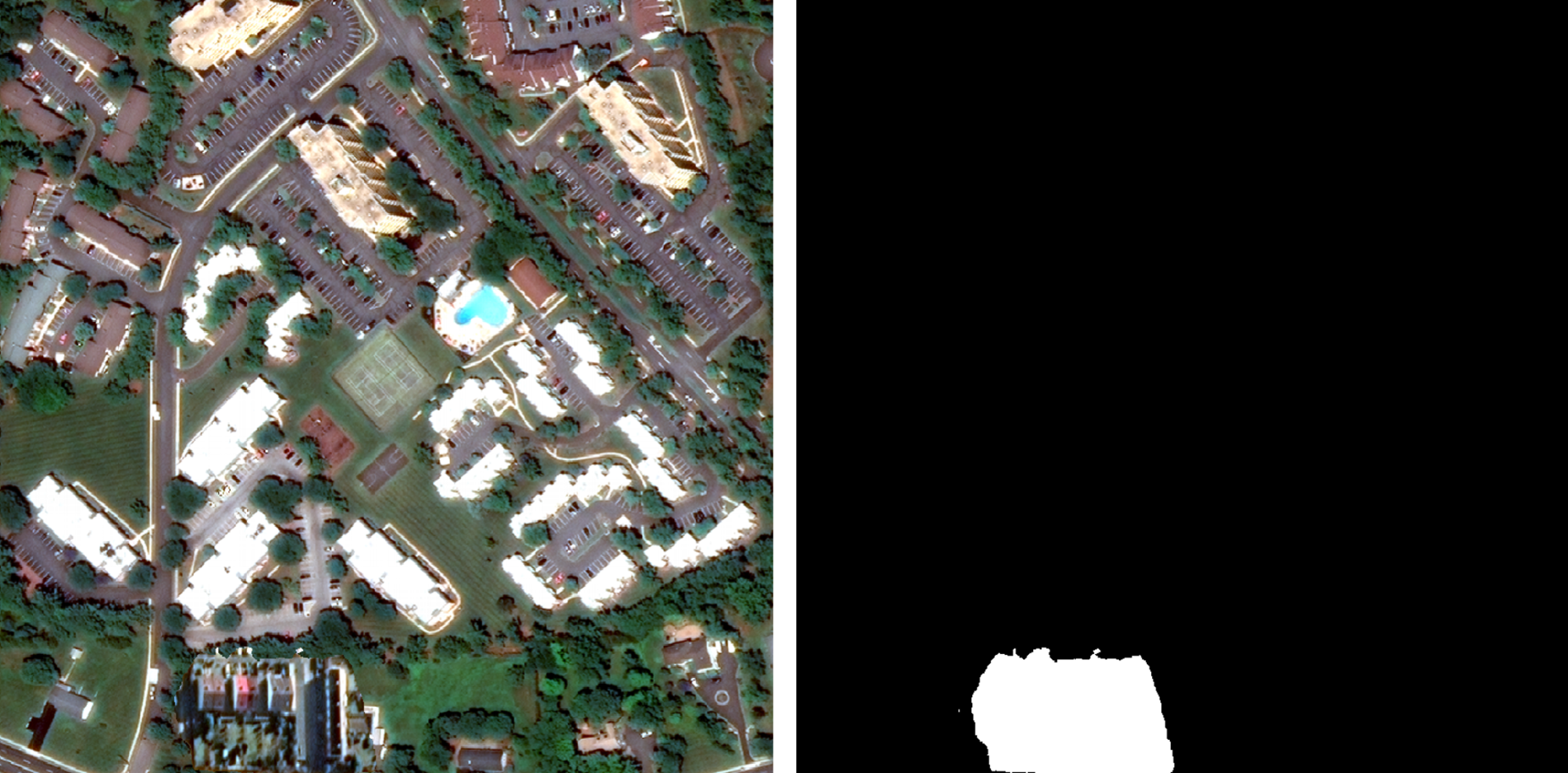}
        \end{subfigure}
    }
    \vspace{0.5em}
    
    \subcaptionbox{Diffusion model inpainted examples. left: green space, center: building, right: airplane.\label{fig:row3}}{
        \begin{subfigure}{0.32\textwidth}
            \centering
            \includegraphics[width=\linewidth]{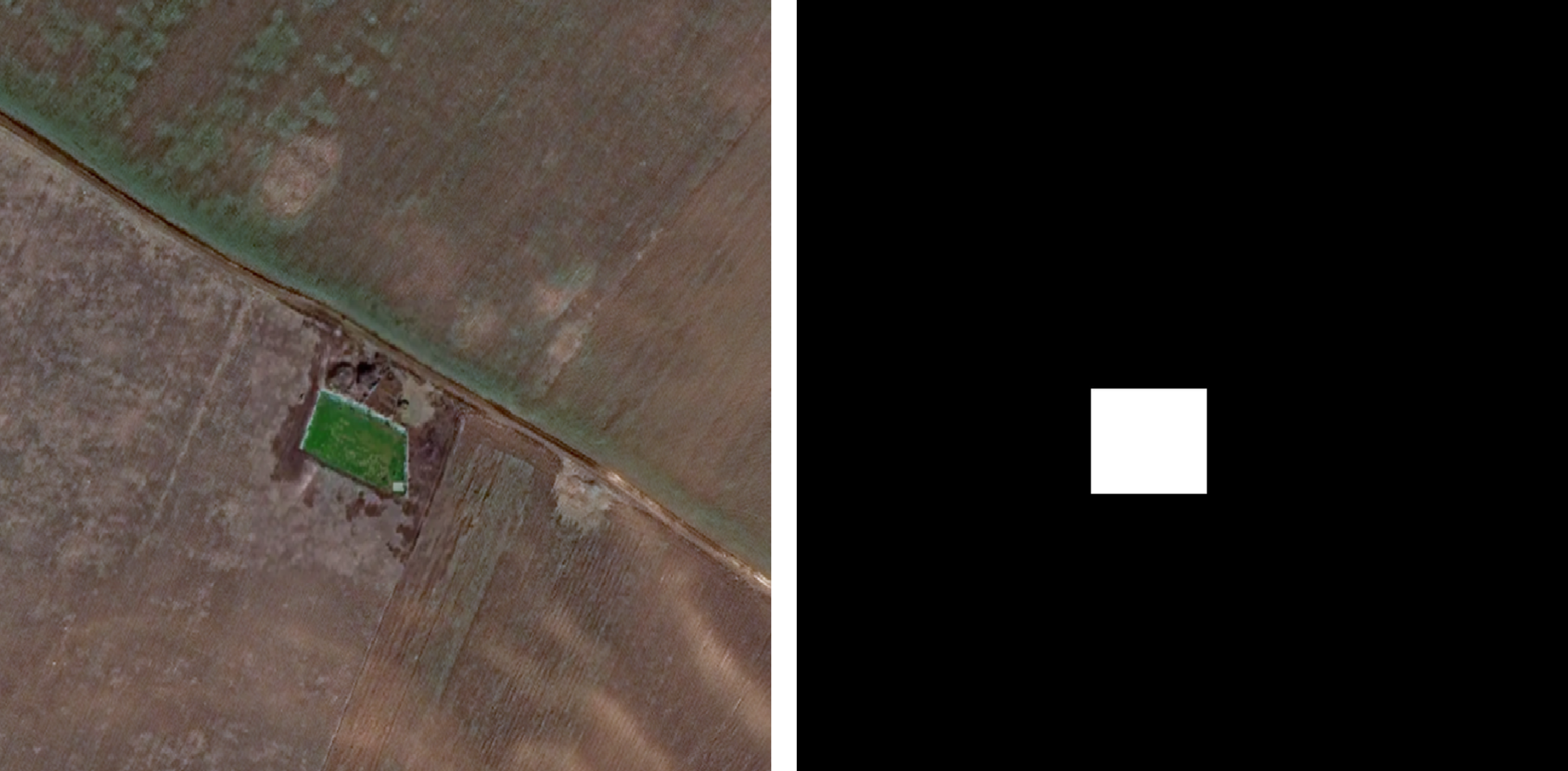}
        \end{subfigure}
        \hfill
        \begin{subfigure}{0.32\textwidth}
            \centering
            \includegraphics[width=\linewidth]{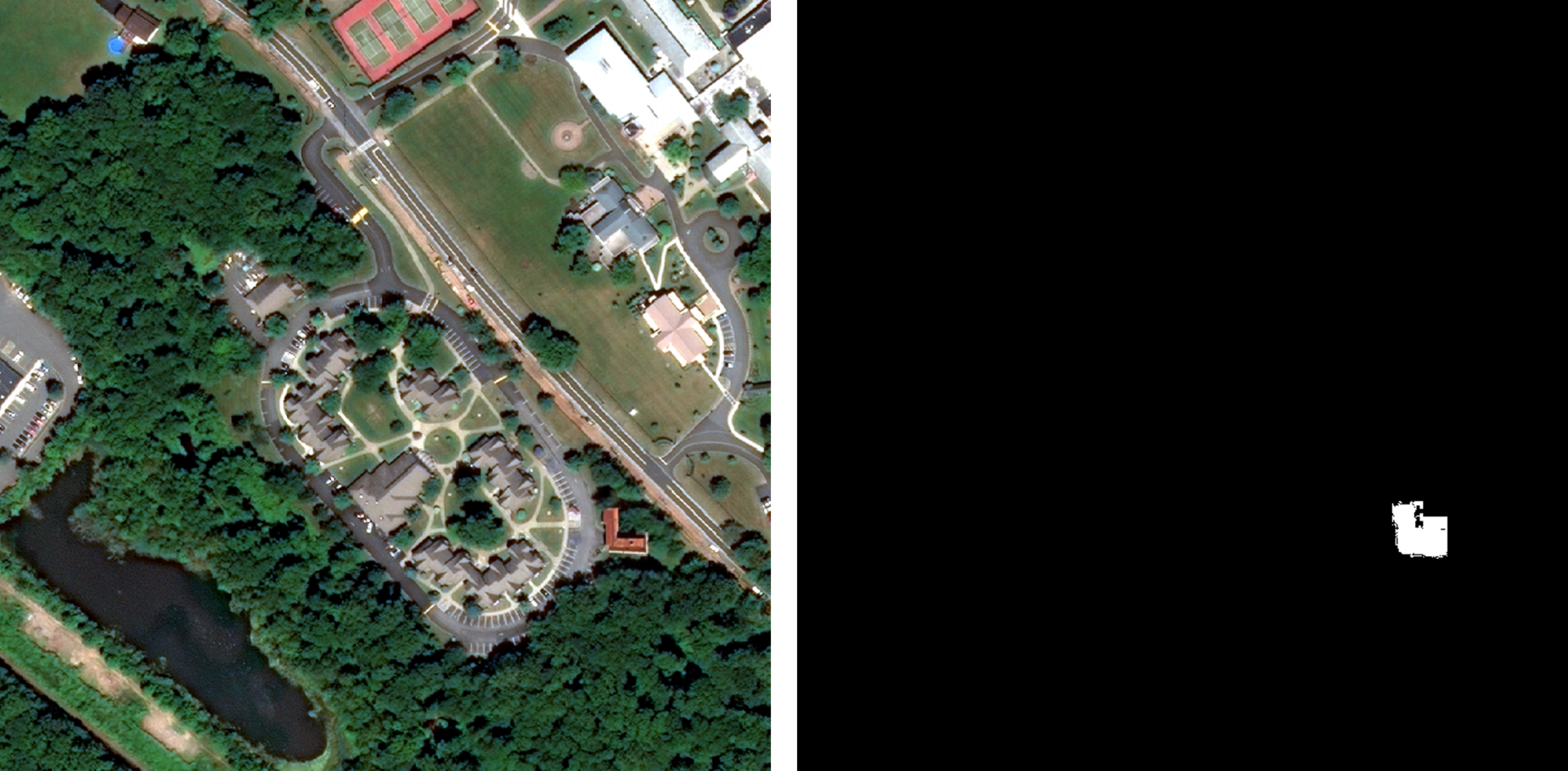}
        \end{subfigure}
        \hfill
        \begin{subfigure}{0.32\textwidth}
            \centering
            \includegraphics[width=\linewidth]{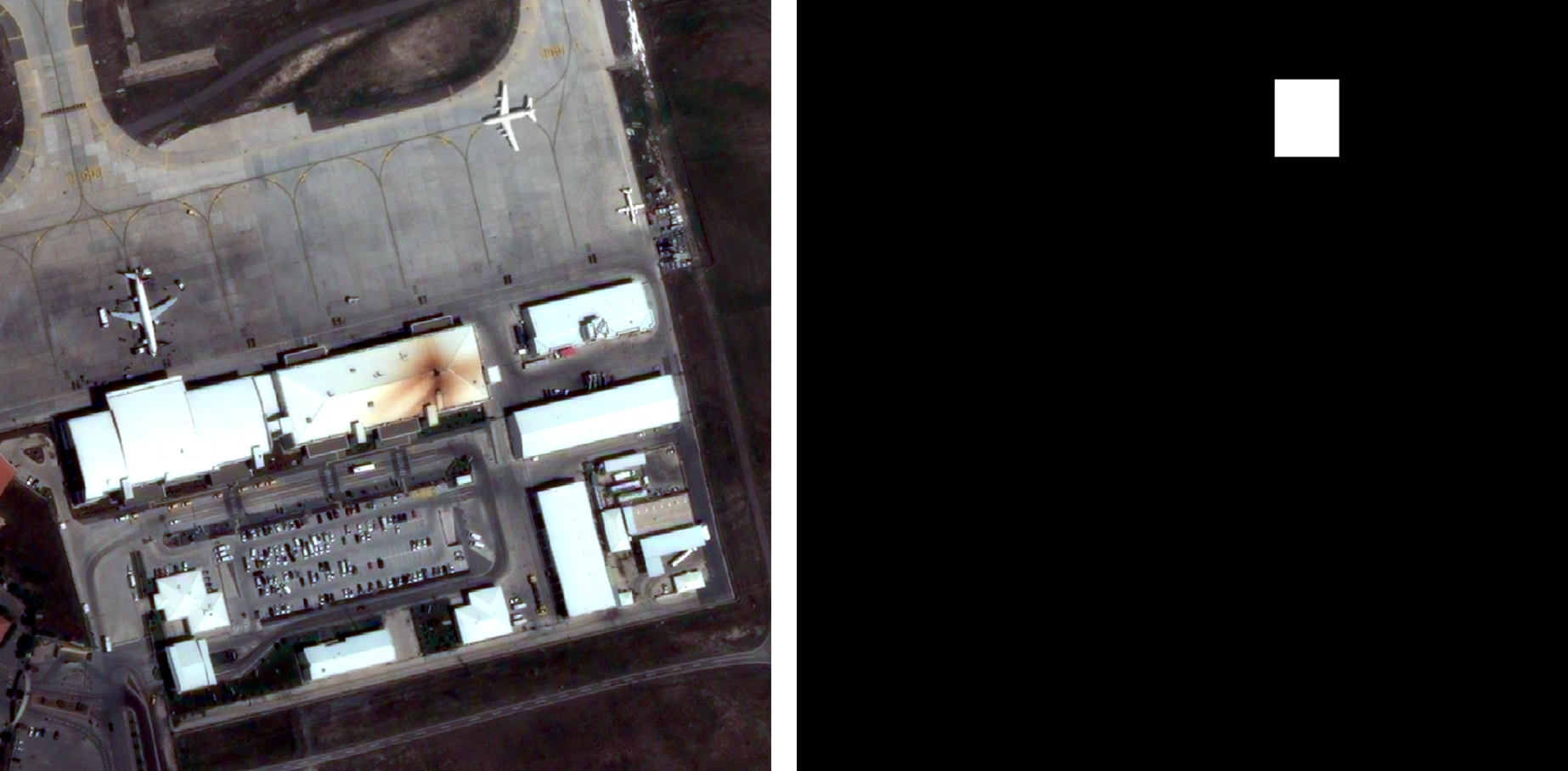}
        \end{subfigure}
    }
\caption{Example fake images and their ground truth masks in the fmow-fake-small dataset.}
\label{fig:fmow-fake-small-examples}
\end{figure*}

Figure~\ref{fig:fmow-fake-small-examples} displays samples in the fmow-fake-small dataset. The fake portion of the dataset includes 10 simple-splice images, 10 object splice images, and 10 inpainted images for a total of 30 fake images. For simple-splices, we include 2 splices at $16\times16px$, 3 at $64\times64px$, 3 at $128\times128px$, and 2 at $256\times256px$. The object splices include the following target objects: \textit{car, pool, building, runway, green space, barren land, helipad}, and \textit{tennis court}. For the inpainting examples, the following reference ``things'' and ``stuff'' are used: \textit{agriculture field, airplane, dirt road, building, track-and-field, barren land}, and \textit{green space}.


\section{Discussion}
In Figure~\ref{fig:datasets_comparison}, we illustrate higher quality fake images with fewer visual artifacts in the fmow-fake-small dataset compared to existing remote sensing deepfake datasets. Our dataset includes diverse image forgeries of different sizes along with ground truth masks for evaluating localization performance, and image acquisition metadata for further analysis. 

However, we recognize several limitations in the current dataset that can be summarized as: i) volume and diversity of generated content, and ii) manual processes for object-splice and inpainting sample curation. First, with only 30 manipulated and 30 real samples, the dataset is quite small in terms of total number of images relative to other available datasets. Furthermore, by only including three manipulation types, and with only one being from a generative model, the diversity of manipulations and AI-generated content is minimal relative to the RSFAKE-1M dataset \cite{tan2025rsfake} but comparable to the other existing remote sensing deepfake datasets (see Table~\ref{tab:related-datasets}). Finally, our method for object splicing is not scalable in its current form and requires human effort to curate splice objects and to place them in logical locations in the base image. Similarly, the inpainting process requires manual effort in defining the mask location for inpainting.

While the required manual effort in the current proposed methodology for dataset construction is a limiting factor in scaling the volume of fake images in the dataset, it guarantees that the dataset includes high-quality deepfakes. As such, we suggest that this dataset is better used as an evaluation dataset rather than for training detectors.




\section{Conclusion}
In this paper, we proposed a new benchmark dataset consisting of copy-paste splices and diffusion model inpainting examples for evaluating remote sensing deepfake localization methods. This dataset addresses the limitation of existing remote sensing deepfake detection datasets by including ground truth masks to assess localization performance and by including deepfakes of high realism. In the future we plan to improve this dataset by increasing the number of manipulated and real images in the dataset, diversifying the pretrained generative models used for synthesizing content, and incorporating fully synthesized images. While this initial dataset is small, it offers a more robust and realistic evaluation dataset for remote sensing deepfake detection algorithms compared to existing datasets.

\section{Acknowledgments}
We acknowledge that this manuscript has been authored by UT-Battelle, LLC under Contract No. DE-AC05-00OR22725 with the U.S. Department of Energy. The United States Government retains and the publisher, by accepting the article for publication, acknowledges that the United States Government retains a non-exclusive, paid-up, irrevocable, world-wide license to publish or reproduce the published form of this manuscript, or allow others to do so, for United States Government purposes. DOE will provide public access to these results of federally sponsored research in accordance with the DOE Public Access Plan (http://energy.gov/downloads/doe-public-access-plan). Research sponsored by the Laboratory Directed Research and Development Program of Oak Ridge National Laboratory, managed by UT-Battelle, LLC, for the U. S. Department of Energy.

\small
\bibliographystyle{IEEEtranN}
\bibliography{references}

\end{document}